\documentclass[10pt,twocolumn,letterpaper]{article}

\usepackage{iccv}
\usepackage{times}
\usepackage{epsfig}
\usepackage{graphicx}
\usepackage{amsmath}
\usepackage{amssymb}
\usepackage{algorithm}
\usepackage[noend]{algpseudocode}

\usepackage[breaklinks=true,bookmarks=false]{hyperref}

\iccvfinalcopy 

\def\iccvPaperID{****} 

\ificcvfinal\fi

\begin{document}

\title{A Recipe for Efficient SBIR Models: Combining Relative Triplet Loss with Batch Normalization and Knowledge Distillation}

\author{Omar Seddati\\
ISIA Lab (UMONS)\\
{\tt\small omar.seddati@umons.ac.be}
\and
Nathan Hubens\\
ISIA Lab (UMONS)\\
{\tt\small nathan.hubens@umons.ac.be}
\and
St\'{e}phane Dupont\\
MAIA (UMONS)\\
{\tt\small st\'{e}phane.dupont@umons.ac.be}
\and
Thierry Dutoit\\
ISIA Lab (UMONS)\\
{\tt\small thierry.dutoit@umons.ac.be}
}
%

%
%
\maketitle
\ificcvfinal\thispagestyle{empty}\fi
%
\begin{abstract}
    Sketch-Based Image Retrieval (SBIR) is a crucial task in multimedia retrieval, where the goal is to retrieve a set of images that match a given sketch query.
    Researchers have already proposed several well-performing solutions for this task, but most focus on enhancing embedding through different approaches such as triplet loss, quadruplet loss, adding data augmentation, and using edge extraction. 
    In this work, we tackle the problem from various angles. We start by examining the training data quality and show some of its limitations. 
    Then, we introduce a Relative Triplet Loss (RTL), an adapted triplet loss to overcome those limitations through loss weighting based on anchors similarity.
    Through a series of experiments, we demonstrate that replacing a triplet loss with RTL outperforms previous state-of-the-art without the need for any data augmentation.
    In addition, we demonstrate why batch normalization is more suited for SBIR embeddings than $l_2$-normalization and show that it improves significantly the performance of our models.
    We further investigate the capacity of models required for the photo and sketch domains and demonstrate that the photo encoder requires higher capacity than the sketch encoder, which validates the hypothesis formulated in \cite{seddati2022towards}.
    Then, we propose a straightforward approach to train small models, such as ShuffleNetv2 \cite{ma2018shufflenet} efficiently with a marginal loss of accuracy through knowledge distillation.
    The same approach used with larger models enabled us to outperform previous state-of-the-art results and achieve a recall of $62.38\%$ at $k = 1$ on The Sketchy Database \cite{sangkloy2016sketchy}.
\end{abstract}

\section{Introduction}
\label{sec:intro}
Sketch-Based Image Retrieval (SBIR) is a fundamental task in multimedia retrieval, where the goal is to retrieve images that match a given sketch query. During the last few decades, the rapid growth in digital media has spurred great interest in multimedia retrieval solutions like SBIR. With the widespread use of touchscreen devices in our daily lives, SBIR solutions have become well-suited for various applications. For instance, an SBIR solution can be integrated into an e-commerce system, where the user draws a sketch to find a specific product. Sketching offers the user a powerful way to convey details beyond the product's category, including global product design and detailed patterns and their actual spatial configuration, which can be difficult to communicate using a text-based query.

Despite the obvious advantages of SBIR solutions, there are several challenges that the computer vision community is still working to overcome. These challenges are related to the abstract nature of sketches, and the gap between natural images and sketches that requires efficient cross-domain features. In addition, the complexity of sketching and sketching quality assessment make new database creation complex and time-consuming. In recent years, researchers have proposed various solutions to address these challenges through more efficient training pipelines, transfer learning, and data augmentation.
In this work, we tackle several aspects of the SBIR problem with the aim of identifying a recipe that would provide practical improvements beneficial to any SBIR application. To achieve our goal, we require a recipe that enables SBIR models to reach high accuracy, and that provides enough flexibility when it comes to choosing the models' size to meet different application requirements (e.g. applications running on devices with limited computing power and storage capacity).

We observe two major limitations of current SBIR solutions, severely hindering their performance. In particular, those observations are (1) Data Unreliability, i.e. there exist multiple instances of the same photo that are so similar that they cannot be differentiated by sketches (which we will refer to as \textit{Ambiguous Samples}); (2) Cross-Domain Misalignment, i.e. there exists a discrepancy in the embedding representation of the photo and sketch models, which should be addressed to ensure a faithful image retrieval. 

In this work, we tackle the Data Unreliability problem by proposing the RTL, a modified version of the triplet loss that takes into account the similarity between anchors (the photos) to weigh the calculated loss. The goal of the RTL is to reduce the impact of \textit{Ambiguous Samples}, which is discussed in more detail in section \ref{sec:id_ambig}. The Cross-Source Domain Misalignment is tackled by introducing batch normalization layers on the embeddings of both models involved. Further explanations regarding this choice will be presented in section \ref{sec:bn}.
%
%
%
%

In \cite{seddati2022towards}, the authors draw attention to the fact that a typical SBIR application has primarily two distinct steps: (1) an \textit{offline} step, where representations for all photos are first extracted and then stored in a database;  (2) an \textit{online} step, where a user draws a sketch used as a query to find the corresponding photo. In the first step, features are extracted for all the photos in a collection (such as product photos on an e-commerce site), and occasional updates are made when new photos are added to the database. This offline part concerns only photos and offers more flexibility in terms of resources (time and computing resources) that can be allocated to this task. The extracted features are then stored in a database intended to be used with a $kNN$-like search approach.
During the second step, resource allocation becomes significantly more critical. On the one hand, we have the user who draws a sketched query and waits for a response. On the other hand, the less resource-intensive the model is, the easier it will be to extract the features of the query sketch locally (on a mobile phone, for example). 
In \cite{seddati2022towards}, the authors have shown that, despite the fact that in all previous works, researchers systematically use the same architecture (e.g. ResNet50) for both modalities (sketch/photo), this was not a mandatory condition. They trained a ResNet18 for sketches and a ResNet34 for photos using the triplet loss and achieved state-of-the-art results. 
Additionally, they hypothesized that efficiently encoding sketches could be achieved with a smaller model compared to encoding photos.  
In this study, we adopt the principle of hybrid architectures to conduct a series of experiments. The results of these experiments show that the latter hypothesis is valid. Additionally, by using a new pipeline that combines the benefits of RTL, batch normalization, and knowledge distillation, we can achieve state-of-the-art results. In our experiments, we replace a ResNet34 with a model as small as a ShuffleNetV2 for sketch encoding without any significant decrease in performance.

To summarize, this paper proposes several contributions to improve SBIR, including:
\begin{itemize}
    \item Examining the training data quality and identifying the issue related to \textit{Ambiguous Samples};
    \item Proposing a Relative Triplet Loss (RTL) to overcome the limitations of traditional triplet loss through loss weighting based on anchor similarity;
    \item Showing that batch normalization is more suited for SBIR embeddings and that it significantly improves the performance of the models;
    \item Investigating and validating the hypothesis made in \cite{seddati2022towards} about the SBIR encoders requirements. We show that indeed photo encoders require higher capacity than sketch encoders;
    \item Proposing a straightforward approach to efficiently train small models, such as ShuffleNetV2, with a marginal loss of accuracy through following a straightforward recipe;
    \item Outperforming the state-of-the-art on the most used large-scale benchmark for SBIR, The Sketchy database, to reach a recall of $62.38\%$ at $k = 1$.
\end{itemize}
Overall, the proposed contributions aim to address the challenges of SBIR and provide a practical recipe for improving its accuracy and adapting to different resource requirements.
%
%
\section{Related Works}
In the field of computer vision, supervised learning using Convolutional Neural Networks (CNNs) has been delivering state-of-the-art results for a few years now \cite{deng2009imagenet,szegedy2015going,he2016deep, cai2022reversible, brock2021high, liu2022convnet, misra2019mish}. The impressive ability of CNNs to extract relevant features directly from pixels without the need for classic feature extraction methods has made them a popular and powerful tool for multiple computer vision tasks. Furthermore, when compared to hand-crafted features (i.e. shallow features), CNN features have been shown to achieve higher performance \cite{gordo2016end, tolias2015particular, seddati2017towards, husain2019remap, kim2018regional, magliani2018accurate} in generating representations for tasks such as content-based image retrieval (CBIR). This same ability of CNN has also significantly improved sketch recognition and SBIR, outperforming previous solutions based on hand-crafted features \cite{laviola2006mathpad2,ouyang2011chemink,cao2013sym,li2013sketch,yesilbek112015svm}.

Several studies have utilized CNNs to develop solutions for sketch-edge map matching. For instance, in \cite{song2019edge,qi2016sketch,seddati2016deepsketch2image,radenovic2018deep}, researchers employed a CNN, initially trained for sketch recognition, to extract features and build an SBIR solution. Their underlying assumption was that photos' edge maps are visually closer to sketches. In a similar vein, Qi et al. \cite{qi2016sketch} used a Siamese CNN architecture for category-level Sketch-Based Image Retrieval (the pose of the object is ignored, only the category matters). The Siamese architecture involves two branches- one for the sketches and the other for the edge maps. During training, the model was fed with sketch-edge map pairs, and a binary label determined if both the sketch and the edge map belonged to the same category. The loss function was then computed, and the model parameters are updated to extract improved representations. Moreover, pair losses were used in a more generalized approach to project inputs into a feature space that minimizes the distance between positive pairs (similar inputs) by a margin of $m_{p}$ while ensuring the distance for negative pairs is larger than a second margin, $m_{n}$. However, using a fixed margin for all pairs is a significant drawback as it fails to account for the variance of (dis)similarity between different pairs. 
To overcome this limitation, researchers have turned to Triplet Loss, which presents inputs as triplets consisting of a reference sample, a positive sample (similar to the reference), and a negative sample (dissimilar to the reference). During training, the model learns to project inputs into a space where a positive example is closer to the reference than a negative one, based on a relative distance measure. This approach allows the model to manage arbitrary feature space distortions and is more suitable for CBIR/SBIR applications, which has garnered significant attention in recent years \cite{song2017deep,chen2022ae,chaudhuri2020crossatnet}.
Bui et al. \cite{bui2016generalisation} investigated in-depth weight-sharing strategies and generalization capabilities of triplet CNNs for SBIR. 
In \cite{seddati2022towards}, the authors conducted a comprehensive study of classic triplet CNN training pipelines in the SBIR context and proposed several avenues for improvement. They highlight the importance of several choices made when building SBIR solutions such as embedding normalization, model sharing, margin selection, batch size, and hard mining selection. 
To overcome the lack of annotated sketches, Bhunia et al. \cite{bhunia2021more} proposed a photo-to-sketch generator using a GAN architecture to synthesize sketches for unlabeled photos. The synthetic sketch-photo pairs were then used to train a triplet CNN. 
In \cite{seddati2022transformers}, the authors proposed a modified sampling pipeline used during training that makes it harder through mini-batches partially filled with samples flipped and a higher number of samples belonging to the same category.
In \cite{zhang2022deformable}, Zhang et al. incorporated a deformable CNN layer to handle sketch variability. 
In addition to the triplet loss, Lin et al. \cite{lin2019tc} experimented with a combination of three loss functions (SoftMax loss, Spherical loss\cite{liu2017sphereface}, and Center loss \cite{wen2016discriminative}). 
Attention modules were also added by some researchers to improve the capturing of fine information \cite{song2017deep,chen2022ae,chaudhuri2020crossatnet,seddati2017triplet}. 
Alternatively, researchers introduced quadruplet networks in \cite{seddati2017quadruplet} to encode semantic information similarly to triplets for local information. 
In a more sophisticated approach, Wang et al. \cite{wang2020deep} proposed a three-stage solution for SBIR, where textual descriptions were used as additional input to the pipeline to reduce the gap between sketches and images. In \cite{sain2021stylemeup}, Sain et al. proposed a cross-modal variational autoencoder to disentangle the semantic content and sketcher style in sketches to build a style-agnostic model. In \cite{seddati2022transformers, ribeiro2020sketchformer}, the authors used a transformer architecture to achieve state-of-the-art results.
In this work, we conduct a comprehensive analysis of various studies to identify the best recipe for creating efficient SBIR solutions. We define an efficient SBIR solution as one that achieves high performance while taking into account the peculiarities of the problem under study, as well as the practical peculiarities that can be leveraged for even greater effectiveness.
%
%
\section{Methodology}
In this section, we present the methodology that we follow to build our recipe and the intuitions behind the different choices we make. We start by introducing the Regional Maximum Activation of Convolutions (RMAC) approach, which we use to measure the similarity between the training photos. Next, we describe our proposed RTL, which overcomes the limitations of the standard triplet loss by incorporating a similarity-based loss weighting mechanism. We then discuss the importance of batch normalization for embeddings, which improves the cross-domain misalignment between the photo and sketch embeddings. Finally, we describe our approach to training efficiently a small model for sketch encoding.
\subsection{Identifying ambiguous samples using RMAC}
\label{sec:id_ambig}
\begin{figure}[t]
    \begin{center}    
        \includegraphics[width=0.8\linewidth]{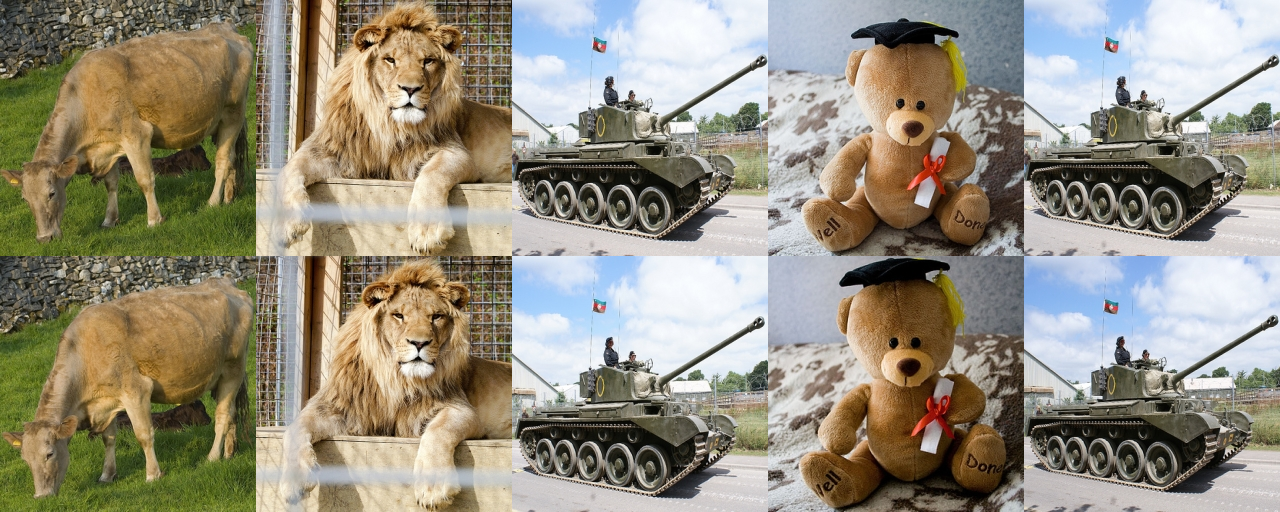}
        \includegraphics[width=0.8\linewidth]{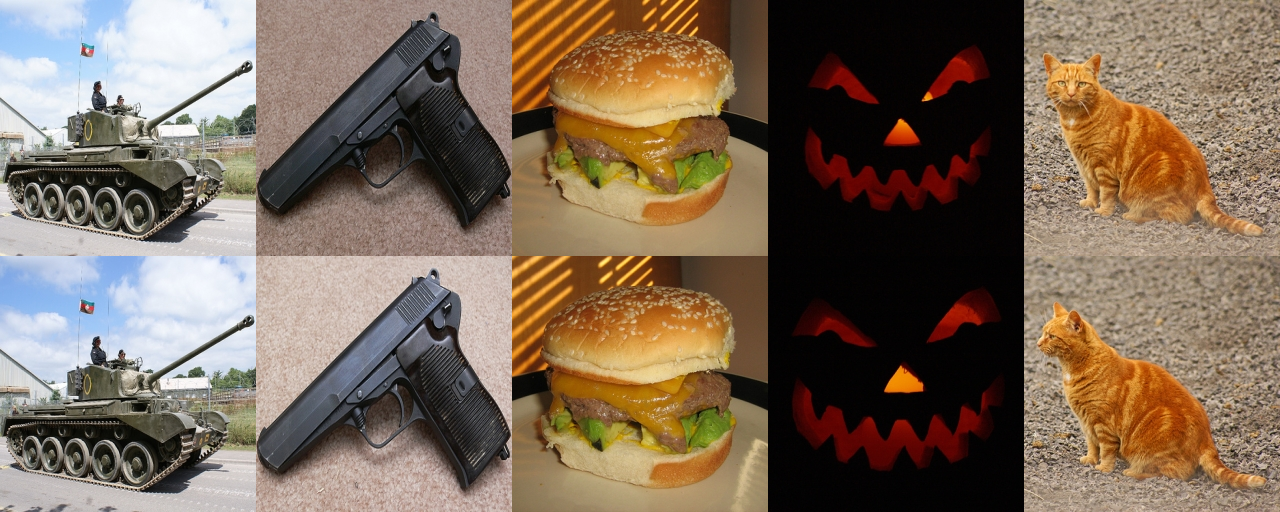}
        \includegraphics[width=0.8\linewidth]{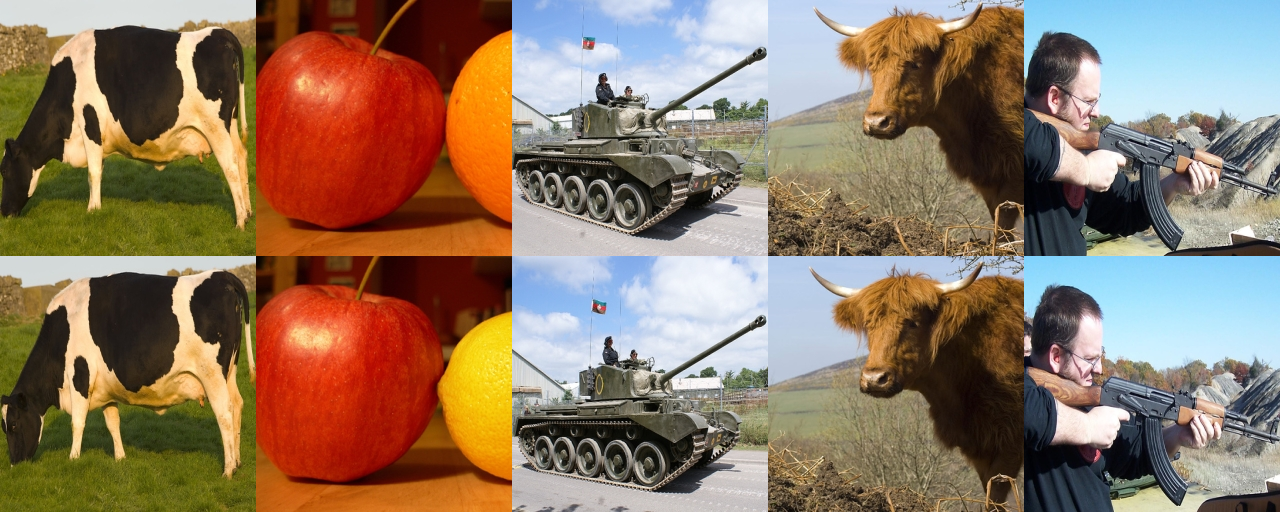}
        \includegraphics[width=0.8\linewidth]{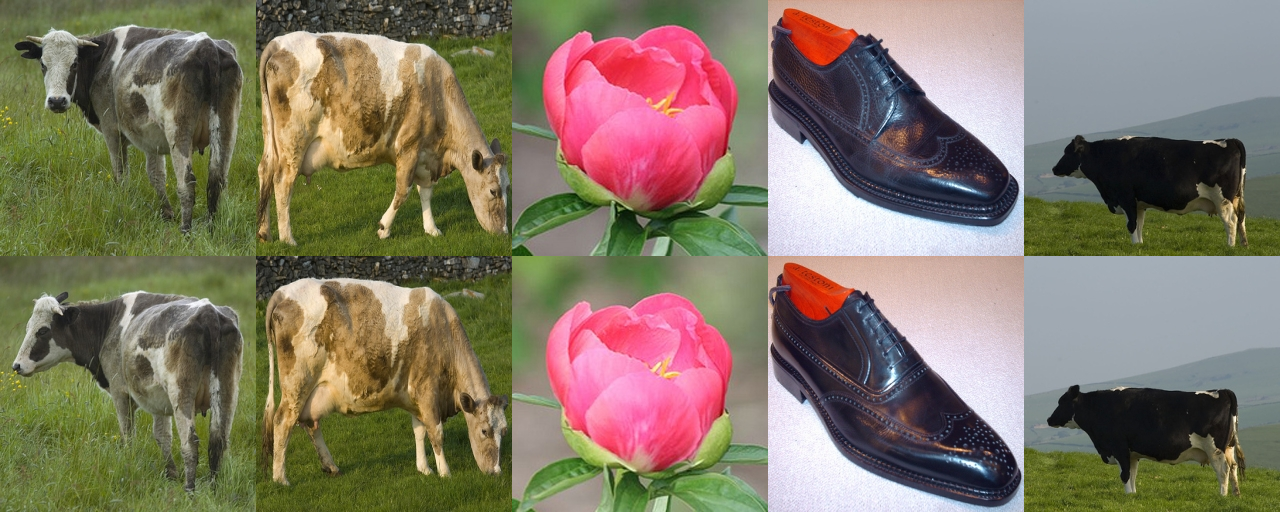}
        \includegraphics[width=0.8\linewidth]{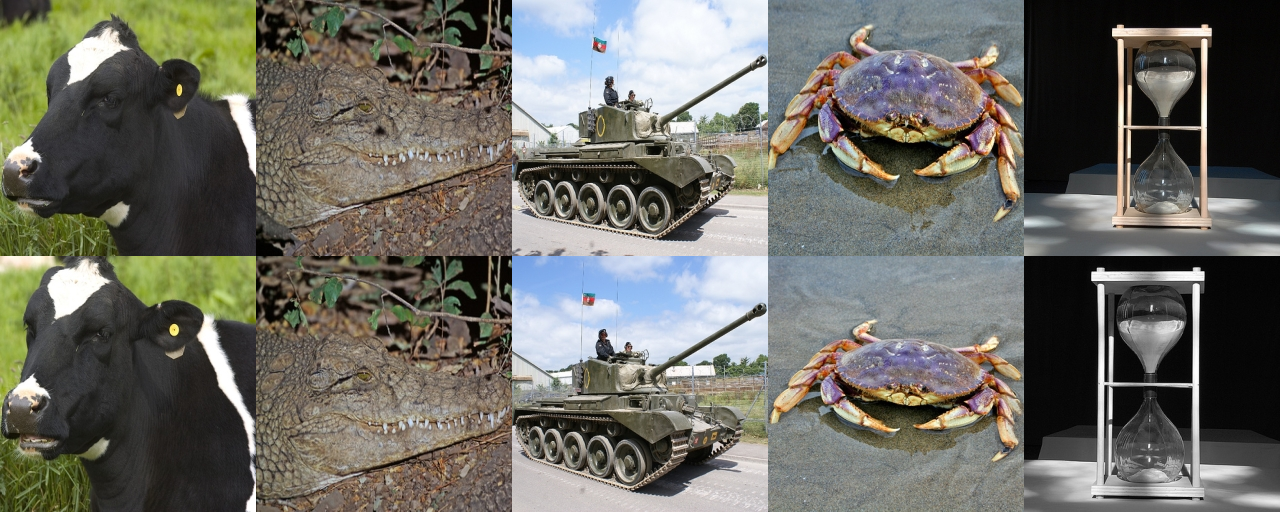}
    \end{center}
       \caption{The top 25 similar pairs of photos retrieved from the training set of the Sketchy benchmark using the RMAC descriptor. In several cases, the images are too similar or even identical to be distinguished using a simple sketch.}
    \label{fig:rmac_db}
\end{figure}

In \cite{tolias2015particular}, the authors demonstrated that a CNN approach can compete with traditional methods on challenging image retrieval benchmarks. To extract features, they discarded the fully connected layers of a pre-trained VGG16 and used the resulting fully convolutional network for feature extraction. For each image input, the output feature maps form a 3D tensor of shape $C\times W\times H$, where $C$ is the number of channels, and $(W,H)$ are the width and height of the feature maps. By representing this tensor as a set of 2D feature maps $\mathcal{X} = {\mathcal{X}{c}}$, $c=1...C$, the Maximum Activations of Convolutions (MAC) can be computed using $\max{x\in \mathcal{X}_{c}} x$ for each $c$.
To compute the RMAC descriptor, Tolias et al. \cite{tolias2015particular} proposed a method to sample a set of square regions $R={R_i}$ within $\mathcal{X}$ using a sliding window approach with a square kernel of width $k_{w}=2\times\min(W,H)/(l+1)$ and stride $60\%\times k_w$ at $L=3$ different scales. Then, for each region, the descriptor $f_{R_{i}}$ is computed using $\sum\limits_{x\in \mathcal R_{i},c} x^\alpha$ with $\alpha = 10$ and normalized using $l_{2}$ normalization, PCA-whitening, followed by an additional normalization. Finally, all the resulting vectors are combined and normalized to obtain the final RMAC descriptor.
Several variants have been proposed \cite{gordo2016end, seddati2017towards, li2017ms, husain2019remap} to build a stronger RMAC descriptor through modifications such as using multi-resolution inputs, features extracted from different layers, normalization, and aggregation. In this work, we adopt some of these modifications: replacing approximate pooling with max pooling, removing PCA-whitening, and using a multi-resolution RMAC descriptor (we use three resolutions for the photos: $S = {384, 512, 768}$). We compute the RMAC descriptor for all the photos on the training set and use the euclidean distance to compare them. Then, we visually checked the top 100 similar pairs of photos to check if the number of \textit{Ambiguous Samples} is significant. As we can see in Figure \ref{fig:rmac_db}, in several cases the images to be discriminated against are too similar or even identical, making it impossible to discriminate them with a simple sketch.
\subsection{Relative Triplet Loss}
\label{sec:rtl}
In this work, we propose RTL, a modified version of triplet loss that aims to incorporate relativity in the loss computation to address the problem described above. In an efficient SBIR solution, the photo encoding must be discriminating enough to meet the margin constraint imposed by the triplet loss. During our experiment, we assume our photo encoder to perform the encoding sufficiently well in order to find a correspondence between a sketch and a photo. Furthermore, as a side-effect of the SBIR training, we assume that our photo encoder is improving at its task during the learning phase. Under these assumptions, the embeddings extracted for photos to compute the triplet loss can be used to measure the similarity, computed using the euclidean distance, between the current mini-batch photos. 
%
%
Let us assume that we have a mini-batch with $bs$ samples (photos). We first compute the similarity matrix $M_{bs\times bs}$ between all the photos in the mini-batch. We then normalize $M_{bs\times bs}$ by dividing it by its maximum value $max(M_{bs\times bs})$ to obtain our weighting matrix $W_{bs\times bs}$. 
To switch from a classic triplet loss to RTL, once we have the triplet loss matrix, we multiply it element-wise by the matrix $W_{bs\times bs}$ before aggregation.
The complete pipeline of our approach is detailed in Algorithm \ref{algo:TRL}.
\begin{algorithm}[H]
    \caption{Relative Triplet Loss}
    \label{algo:TRL}
    \begin{algorithmic}[1]
        \Require
        \State Batch size: $bs$
        \State Batch of photos: $P$, $P_i$ where $i=0...bs$
        \State Batch of sketches: $S$, $S_i$ where $i=0...bs$ and $S_i$ is a sketch matching the photo $P_i$
        \State Margin: $m$
        \State Photos embedding function: $f_p(\cdot)$
        \State Sketches embedding function: $f_s(\cdot)$
        \State Distance function: $D(\cdot,\cdot)$
        \State Rectified linear unit: $ReLU(\cdot)$
        \State Identity matrix: $I_{bs}$
        \Ensure
        RTL loss: $L_{RTL}$
        \State Mini-batch photos embeddings: $P\_embs = f_p(P)$
        \State Mini-batch sketches embeddings: $S\_embs = f_s(S)$
        \State Distance between the anchors and positive samples: $d_{a,p} = D(P\_embs_i, S\_embs_j)$ with $i = j$
        \State Distance between the anchors and negative samples: $d_{a,n} = D(P\_embs_i, S\_embs_j)$ with $i \ne j$
        \State We expand $d_{a,p}$ and compute the triplet loss matrix: $TL_{matrix} = ReLU(d_{a,p} - d_{a,n} + m)$
        \State Then, we compute the weighting matrix: $W_{bs\times bs} = D(P\_embs, P\_embs)/max(D(P\_embs, P\_embs))$
        \State We then compute the RTL matrix: $RTL_{matrix} = TL_{matrix} \times (1 - I_{bs}) \times W_{bs\times bs}$        
        \State Compute the sum of $RTL_{matrix}$ to obtain the final $RTL$ loss: $L_{RTL}=sum(RTL_{matrix})$
    \end{algorithmic}
\end{algorithm}
\subsection{Batch normalization for embeddings}
\label{sec:bn}
The internal feature distributions of neural networks are highly dependent on the domain that they are operating on, which makes it difficult to directly compare distributions in a cross-source setting.
To alleviate such a distribution shift and encourage a better distribution alignment between our two models, we propose to draw inspiration from the batch normalization technique and to normalize the output activations of each domain model via domain-specific normalization statistics.
Because it is less sensitive to outliers, batch normalization better preserves the representation range of the embeddings than other commonly used normalization schemes such as $l_2$-normalization. As a result, we find that models using batch normalization on their embeddings have well-behaved training dynamics and reach better performances. We hypothesize that thanks to its learnable parameters, batch normalization allows embeddings originating from the sketch model and those from the photo model to be represented in comparable distributions.
%
%
%
%
%
%
%
\subsection{Training a small model for sketches encoding}
\label{sec:train_small}
As explained in the introduction, in order to reduce the resources needed for the online part of an SBIR solution, we can use smaller models to encode sketches. 
In our case, we have opted for ShuffleNetV2 (we use the pre-trained shufflenet\_v2\_x1\_0 from torchvision), a state-of-the-art model that achieves high accuracy with low computation costs. Its tradeoff between accuracy and low computation costs makes it an adequate candidate for SBIR applications. ShuffleNetV2 was designed to meet the needs of mobile devices (limited computing power and storage capacity) and real-time applications (fast inference speed).

 Early experiments conducted on ShuffleNetV2 as a sketch encoder have revealed struggles in convergence, leading to a significant decrease in performance when compared to larger models such as ResNet34. This phenomenon can be explained by the drastically low number of parameters (ShuffleNetV2 has almost 20 times fewer parameters than ResNet34), making it difficult for such a small model to capture the non-negligible complexity of the cross-domain inputs.
 In order to overcome this last hurdle, we came up with the idea of using knowledge distillation to transfer knowledge from a large model pre-trained to encode sketches and that has proven its effectiveness, to a smaller model (ShuffleNetV2 in our case). In this manner, we can circumvent the complexity related to the cross-modality nature of the training and focus more on the validity of the initial hypothesis (small models are enough for sketch encoding). 
%
Knowledge Distillation techniques work by transferring the knowledge of a large and powerful model, the teacher, to a smaller and simpler one, the student, by having the student model regress the output of the teacher. Such a method usually leads to students having better generalization capabilities since the teacher's output implicitly encodes more information about the similarity between training samples and their distribution than hard labels. 

In this work, we propose to apply such a training strategy to our models. In particular, we use a response-based knowledge distillation technique, where the student learns to mimic the output embeddings of a teacher. In that regard, several learning objectives have been addressed, providing different convergence abilities to the student. The respective output embeddings of the teacher and the student have been compared according to (1) Mean-Squared Error; (2) Huber Loss; (3) A combination of Mean-Squared Error and Mean-Absolute Error.

We also explore variants of the traditional knowledge distillation techniques, by using students of comparable or even larger capacities than the teacher. Such an alteration has been shown to lead to student models learning a model ensemble jointly with regular knowledge distillation and to lead to a better-performing student \cite{allen-zhu2023towards}.
%
%
%
%
\section{Experiments}
In this section, we detail the different experiments conducted for this study. 
For the whole study, we utilize The Sketchy benchmark \cite{sangkloy2016sketchy}, a large-scale comprehensive collection designed specifically for SBIR. This benchmark comprises 75,471 sketches for 12,500 unique objects across 125 categories (the benchmark contains 100 photos per category).
To create this dataset, crowd workers were instructed to sketch various photographic objects, resulting in a diverse range of sketch styles and interpretations. For each photo, there are at least five sketches from different workers to ensure a robust set of fine-grained associations between sketches and photos.
To ensure consistency, the authors provide a series of guidelines to follow, including a test set list to split data into a training and test set. Specifically, 90\% of the data are used for training, and the remaining 10\% are used at test time. We follow these guidelines to ensure a fair and reliable evaluation of our models' performance.
\subsection{RTL and batch Normalization for better embeddings}
At the beginning of our experiments, we follow the pipeline proposed in \cite{seddati2022transformers} with some minor modifications. In particular, we do not use a ResNet50 or a Transformer model but use a ResNet18 and a ResNet34 instead. As in \cite{seddati2022transformers} we use pre-trained versions of these models (trained on ImageNet \cite{deng2009imagenet}) provided by the torchvision library. We also use the output of the last pooling layer (adaptive average pooling) to extract the embeddings (without applying $l_{2}$-normalization).
We use two distinct encoders for sketches and photos. We train our models for $200$ epochs. We set the learning rate to $lr = 10^{-4}$ for the first $100$ epochs and we change it to $lr = 10^{-6}$ for the second $100$ epochs. The batch-size $bs$ and the margin $m$ are kept constant for this study, we use $bs = 256$ (instead of $bs = 128$ in \cite{seddati2022transformers}) and $m = 3$.

In Table \ref{table:res_rtl_bn}, we compare our results with equivalent models from \cite{seddati2022transformers} that we consider as baselines.
We can see that replacing triplet loss with RTL and adding a batch normalization layer, both bring significant improvements.
\begin{table}
    \begin{center}
        \begin{tabular}{|l|c|}
        \hline
        Model & $Recall @1\%$ \\
        \hline\hline
        $R18$ \cite{seddati2022transformers} & 52.98 \\
        $R18_{RTL}$ & 55.27 \\
        $R18_{RTL+BN}$ & 57.20 \\
        \hline\hline
        $R34$ \cite{seddati2022transformers} & 56.10 \\
        $R34_{RTL}$ & 58.50 \\
        $R34_{RTL+BN}$ & 59.99 \\
        \hline
        \end{tabular}
    \end{center}
    \caption{Our results achieved on The Sketchy Database with RTL and batch normalization ($BN$) compared to \cite{seddati2022transformers}.}
    \label{table:res_rtl_bn}
\end{table}
\subsection{Training efficiently a small encoder for sketches}
\subsubsection{Training with RTL and batch normalization}
We used our best photo encoder ($R34_{RTL+BN}$) from previous experiments (we freeze all the layers of the photo encoder, including batch normalization parameters) and a $ShuffleNetV2_{sketches}$ for sketch encoding (we replaced the classification layer with a fully connected layer with 512 outputs to reduce the number of channels, followed by a batch-normalization layer). We followed the same training pipeline as before with RTL and batch normalization. We trained the model for 200 epochs and noticed the performance decreased by more than $8\%$. To verify if such a decrease is an indicator of a model limitation or that the model is struggling to converge, we train for longer. As shown in Table \ref{table:shuff_r34_beg}, we can see that after relatively long training, the model ends up reaching results closer to those achieved with $R18_{RTL+BN}$ (Table \ref{table:res_rtl_bn}).

However, despite these satisfactory results, we still have nearly $4\%$ decrease in accuracy compared to an $R34_{sketches}$ and a training strategy that starts to show some weaknesses that should not be ignored in our quest for a straightforward recipe for efficient SBIR solutions.

\begin{table}
    \begin{center}
        \begin{tabular}{|l|c|c|}
        \hline
        Model & Epochs & $Recall @1\%$ \\
        \hline\hline
        $ShuffleNetV2_{sketches}$ & 200 & 51.96 \\
        $ShuffleNetV2_{sketches}$ & 300 & 52.81 \\
        $ShuffleNetV2_{sketches}$ & 500 & 54.19 \\
        $ShuffleNetV2_{sketches}$ & 600 & 54.86 \\
        $ShuffleNetV2_{sketches}$ & 700 & 55.66 \\
        $ShuffleNetV2_{sketches}$ & 800 & 55.97 \\
        $ShuffleNetV2_{sketches}$ & 900 & 56.01 \\
        \hline
        \end{tabular}
    \end{center}
    \caption{Our results achieved on The Sketchy Database with $ShuffleNetV2_{sketches}$ and $R34_{photos}$.}
    \label{table:shuff_r34_beg}
\end{table}
\subsubsection{Knowledge distillation }
As mentioned in section \ref{sec:train_small}, knowledge distillation offers an attractive solution to avoid dealing directly with the complicated nature of cross-modality training. In addition, it provides a good solution to test the initial hypothesis about the sketch encoder size. During the training, we noticed that with this approach the training became fast (it needs less than 200 epochs to converge) and smooth (the evolution is stable).
In Table \ref{table:shuff_r34_distil} (we removed the $_{sketches}$ subscript for better readability), we report our results for the experiments with $R34_{sketches}$ knowledge transfer to $ShuffleNetV2_{sketches}$ (pre-trained on ImageNet). We can see that this approach enabled us to almost reach our initial goal, at this point we are only $1.5\%$ far away.

\begin{table}
    \begin{center}
        \begin{tabular}{|l|c|}
        \hline
        Model & $Recall @1\%$ \\
        \hline\hline
        $ShuffleNetV2_{KL}$ & 53.3 \\
        $ShuffleNetV2_{KL+SM}$ & 56.6 \\
        $ShuffleNetV2_{MSE}$ & 57.71 \\
        $ShuffleNetV2_{MSE+MAE}$ & 58.31 \\
        $ShuffleNetV2_{Huber}$ & 58.5 \\
        \hline
        \end{tabular}
    \end{center}
    \caption{Our results achieved on The Sketchy Database after knowledge distillation from $R34_{sketches}$ to $ShuffleNetV2_{sketches}$. The $R34_{photos}$ is used as the photo encoder.}
    \label{table:shuff_r34_distil}
\end{table}
\subsubsection{Double guidance for finetuning after knowledge distillation}
\begin{figure}[t]
    \begin{center}    
        \includegraphics[width=1.0\linewidth]{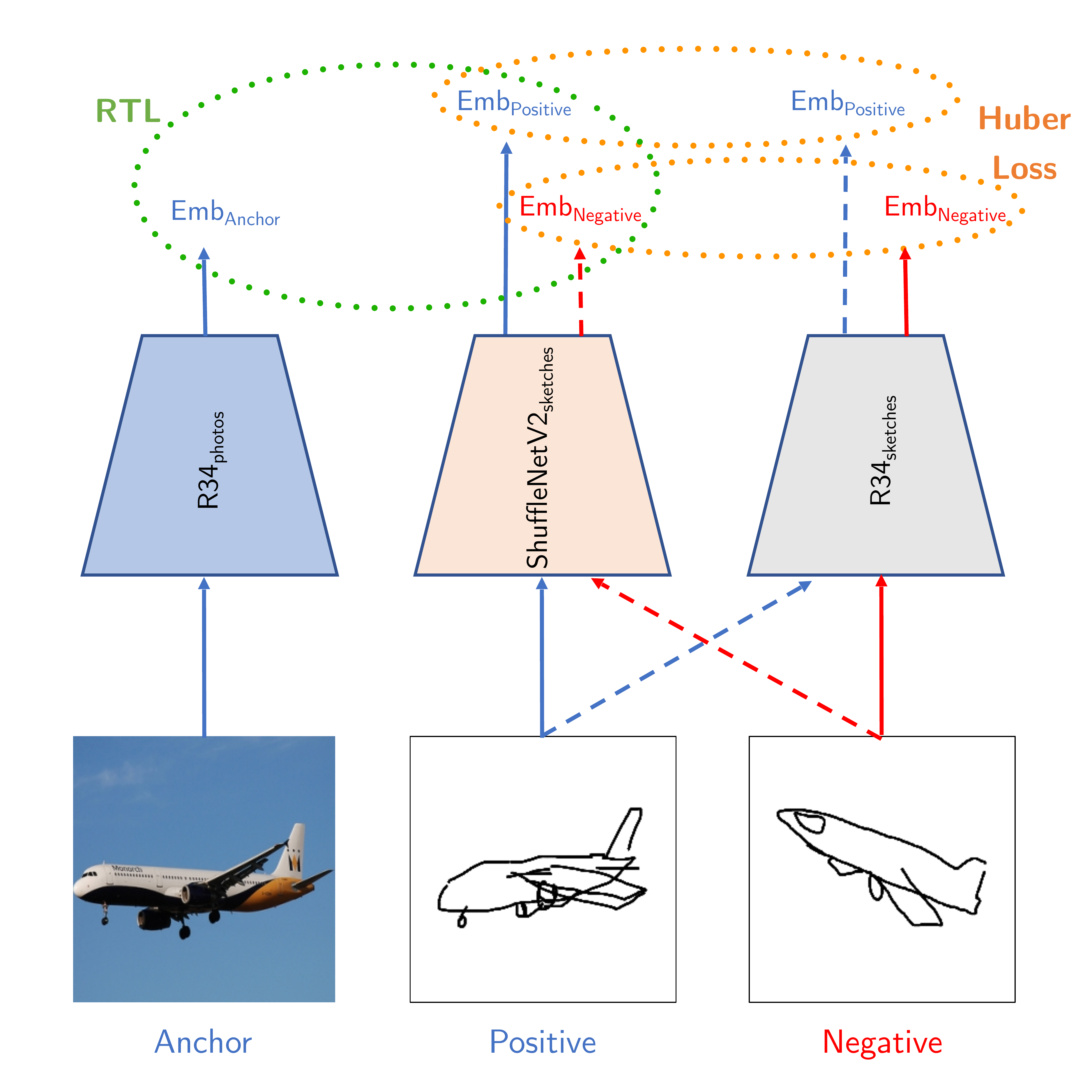}
    \end{center}
       \caption{The proposed double guidance pipeline for an efficient finetuning after knowledge distillation. In this novel pipeline, an additional branch with a sketch encoder teacher is used to guide the student model with Huber Loss.}
    \label{fig:double_guid}
\end{figure}
An intuitive and obvious next step after the success achieved with knowledge distillation was to finetune the new efficient $ShuffleNetV2_{sketches}$ with RTL and $R34_{photos}$. To do so, we started following the previously used pipeline. But unfortunately, even after extensive hyperparameter tuning, the accuracy continued to drop with training. We started to believe that a partial ability acquired during the knowledge distillation phase, is not a requirement for the triplet loss constraint. If this assumption is valid, then it is possible that the model loses it during the finetuning.

In order to alleviate that deficiency, we propose a double guidance pipeline for finetuning after knowledge distillation. In this new pipeline, we use both, the $R34_{photos}$ and the $R34_{sketches}$ at the same time to train our $ShuffleNetV2_{sketches}$. The parameters of $R34_{photos}$ and $R34_{sketches}$ are not updated, the models are only used to guide the $ShuffleNetV2_{sketches}$. 
While the latter learns to extract embeddings that respect the RTL constraint with those of $R34_{photos}$, at the same time, it also learns to mimic the embeddings generated with $R34_{sketches}$ for sketches thanks to the additional Huber loss as shown in Figure \ref{fig:double_guid}. 
This new pipeline enabled us to gain an additional $0.6\%$ to reach a $recall@1 = 59.1$. 
In Table \ref{table:shuff_r34_distil_ft} (we removed the $_{sketches}$ subscript for better readability), we report our result for double guidance finetuning, in addition to results of knowledge distillation experiments with models larger than $ShuffleNetV2$ for sketch encoding, and even larger than the teacher model $R34_{sketches}$. 
As we can see, comparable results were achieved with models of significantly different sizes.
We assume that these results are sufficient proof of the validity of the initial hypothesis for sketch encoding. Following our pipeline, a small sketch encoder can be used for SBIR applications with a marginal loss of accuracy.
\begin{table}
    \begin{center}
        \begin{tabular}{|l|c|}
        \hline
        Model & $Recall @1\%$ \\
        \hline\hline
        $ShuffleNetV2_{Huber}$ & 58.5 \\
        $ShuffleNetV2_{Huber+DG}$ & 59.1 \\
        $R18_{Huber}$ & 59.8 \\
        $R50_{Huber}$ & 59.89 \\
        $R101_{Huber}$ & 60.24 \\
        $R152_{Huber}$ & 59.7 \\
        \hline
        \end{tabular}
    \end{center}
    \caption{Our results achieved on The Sketchy Database after knowledge distillation from $R34_{sketches}$ to multiple models. For this experiment, the $R34_{photos}$ is used as the photo encoder ($DG$ is used to indicate that a double guidance finetuning pipeline was used after knowledge distillation).}
    \label{table:shuff_r34_distil_ft}
\end{table}
\subsection{Training Large Encoders for Photos}
The second part of the initial assumption was about the necessity of relatively big models to encode photos efficiently. We proceed in a similar manner as before to check the validity of this hypothesis.
This time, we use the $R34_{photos}$ for knowledge distillation, and we analyze the performance evolution. In Table \ref{table:ph_distil} (we removed the $_{photos}$ subscript for better readability), we summarize the results of our experiments. We can conclude that the size of the photo encoder matters. We can also notice that a large student is even able to surpass the teacher with little effort. Unlike training with triplet loss, knowledge distillation shows more stability during training, in addition, hyperparameter tuning is easier and less time-consuming.
\begin{table}
    \begin{center}
        \begin{tabular}{|l|c|}
        \hline
        Model & $Recall @1\%$ \\
        \hline\hline
        $ShuffleNetV2_{Huber}$ & 54.31 \\
        $R18_{Huber}$ & 56.62 \\
        $R50_{Huber}$ & 60.7 \\
        $R101_{Huber}$ & 61.98 \\
        $R152_{Huber}$ & 62.38 \\
        \hline
        \end{tabular}
    \end{center}
    \caption{Our results on The Sketchy Database after knowledge distillation from $R34_{photos}$ to multiple models. For this experiment, the $R34_{sketches}$ is used as the sketch encoder.}
    \label{table:ph_distil}
\end{table}
\subsection{Combining a small sketch encoder with large photo encoders}
In Table \ref{table:sk_small_ph_big} (we removed the $_{photos}$ subscript for better readability), we report the results obtained when combining our $ShuffleNetV2_{Huber+DG}$ and different models larger than the teacher model ($R34$). We notice that we have been able to surpass even the initial performance achieved with a ResNet34 used for both encoders. In addition, the proposed recipe offers attractive flexibility that enables the development of SBIR solutions with multiple backbones meeting different requirements.
\begin{table}
    \begin{center}
        \begin{tabular}{|l|c|}
        \hline
        Photo encoder & $Recall @1\%$ \\
        \hline\hline
        $R34$ & 59.1 \\
        $R50_{Huber}$ & 59.18 \\
        $R101_{Huber}$ & 60.88 \\
        $R152_{Huber}$ & 61.45 \\
        \hline
        \end{tabular}
    \end{center}
    \caption{Our results on The Sketchy Database after knowledge distillation and double guidance of the sketch encoder ($ShuffleNetV2_{Huber+DG}$) tested with multiple photo encoders.}
    \label{table:sk_small_ph_big}
\end{table}
\subsection{Comparison with state-of-the-art methods}
In this section, we compare some of our study results with those of previous research on The Sketchy benchmark. These results are reported in Table \ref{table:bench}. 
As can be seen in this table, if we compare our results with others with the same architecture (e.g. ResNet18, ResNet34, ResNet50), we notice that using RTL and batch normalization alone bring a significant improvement. And that they surpass even ResNet18$_{2\times 2}$ and ResNet34$_{2\times 2}$ proposed in \cite{seddati2022transformers}, where the last average pooling was modified to reduce the spatial resolution to $2 \times 2$, which increases four times the embedding size. In addition, our largest distilled photo encoder $R152_{Huber}$, when used with the sketch encoder $R34_{RTL+BN}$, they achieve comparable results to those of the double vision transformer solution proposed in \cite{seddati2022transformers}. And the latter is the only solution that we found in SBIR literature to surpass the results achieved by our hybrid solutions (different architectures for the sketch encoder and photo encoder), even when a $ShuffleNetV2$ is used as sketch encoder.
\begin{table}
    \begin{center}
        \begin{tabular}{|l|c|}
        \hline
        Model & $Recall @1 (\%)$ \\
        \hline\hline
        Chance \cite{sangkloy2016sketchy}&0.01\\
        Sketch me that shoe \cite{yu2016sketch}&25.87\\
        Siamese Network \cite{sangkloy2016sketchy}&27.36\\
        Triplet Network \cite{sangkloy2016sketchy}&37.10\\
        Quadruplet\_MT \cite{seddati2017quadruplet}&38.21\\
        DCCRM(S+I) \cite{wang2020deep}&40.16\\
        DeepTCNet \cite{lu2018instance}&40.81\\
        Triplet attention \cite{seddati2017triplet}&41.66\\
        Quadruplet\_MT\_v2 \cite{seddati2017quadruplet}&42.16\\
        LA \cite{xu2021dla}& 43.1\\
        DCCRM(S+I+D) \cite{wang2020deep}&46.20\\
        Human \cite{sangkloy2016sketchy}&54.27\\
        ResNet18 \cite{chen2022ae}& 45.95\\
        DCCRM \cite{wang2020deep}& 46.20\\
        ResNet50 \cite{chen2022ae}& 52.19\\
        ResNet18 \cite{seddati2022towards}& 52.75\\
        ResNet18 \cite{seddati2022transformers}& 53.61\\
        ResNet101 \cite{chen2022ae}& 54.59\\
        DLA \cite{xu2021dla}& 54.9\\
        ResNet18$_{2\times 2}$ \cite{seddati2022transformers}& 55.10\\
        \bf Our $R18_{RTL}$ & \bf 55.27 \\
        ResNet50  \cite{seddati2022transformers}& 56.29\\
        \bf Our $R18_{RTL+BN}$ & \bf 57.20 \\
        MLRM \cite{ling2022multi}& 57.20\\
        ResNet34  \cite{seddati2022transformers}& 57.43\\
        ResNet34$_{2\times 2}$  \cite{seddati2022transformers}& 58.23\\
        ResNet50$_{2\times 2}$  \cite{seddati2022transformers}& 58.37\\
        \bf Our $R34_{RTL}$ & \bf 58.50 \\
        \bf Our $ShuffleNetV2_{Huber+DG}$ & \bf 59.1 \\
        \bf Our $R34_{RTL+BN}$ & \bf 59.99 \\
        VT  \cite{seddati2022transformers}&62.25\\
        \bf Our $R152_{Huber}$ & \bf 62.38 \\
        \hline
        \end{tabular}
    \end{center}
    \caption{Our main results compared to state-of-the-art solutions on The Sketchy Database. We can observe that our training recipe brings significant improvements for models with the same architecture.}
    \label{table:bench}
\end{table}
%
\section{Conclusion}
In this paper, we have presented a comprehensive study on improving SBIR solutions by tackling some of its major limitations. Starting with pointing out and demonstrating the existence of an issue with data reliability that has been largely ignored. 
To address this problem, we have proposed a Relative Triplet Loss (RTL), a modified version of the triplet loss, that takes into account the similarity between anchors to relatively adapt the computed loss. 
We have also shown that batch normalization is more suitable for SBIR embeddings compared to adding an $l_2$-normalization layer, and it significantly improves the performance of our models. 
Furthermore, we have investigated the capacity of models required for the photo and sketch domains and demonstrated that the photo encoder requires a higher capacity than the sketch encoder. 
Additionally, we have proposed a straightforward recipe based on knowledge distillation to efficiently train small models and even reach higher accuracy with larger ones. 
Our experimental results demonstrate that our proposed method outperforms previous state-of-the-art results and provides a strong pipeline for building more efficient SBIR solutions.
Overall, our work provides a practical recipe for improving both the performance and the efficiency of SBIR systems, which can benefit a wide range of applications, including e-commerce systems.
{\small

}

\end{document}